\documentclass{article}
\usepackage{hyperref}

\usepackage[preprint]{icml2026}
\usepackage{microtype}
\usepackage{graphicx}
\usepackage{booktabs}
\usepackage{array}
\usepackage{amsmath}
\usepackage{amssymb}
\usepackage[shortlabels]{enumitem}
\usepackage{eurosym}
\usepackage[T1]{fontenc}
\usepackage[utf8]{inputenc}
\usepackage{newunicodechar}
\newunicodechar{€}{\euro{}}
\newcommand{\gtfengrelease}{We release gtfeng, an open-source Python package implementing the framework.}
\newcommand{\gtfengpackage}{The open-source Python package gtfeng implements the evaluation procedure and the standard report.}

\icmltitlerunning{GT-FENG: Factual Evaluation and Normative Grounding}
\hypersetup{pdfsubject={Preprint},pdfcreator={},pdfproducer={}}
\begin{document}
\twocolumn[
  \icmltitle{GT-FENG: General Theory of Factual Evaluation and\\ Normative Grounding in Language Models}
  \begin{icmlauthorlist}
    \icmlauthor{Jiayu Feng}{hsph}
  \end{icmlauthorlist}
  \icmlaffiliation{hsph}{Harvard T.H. Chan School of Public Health, Boston, MA, USA}
  \icmlcorrespondingauthor{Jiayu Feng}{jiayufeng@hsph.harvard.edu}
  \icmlkeywords{factuality evaluation, normative facts, error attribution, legal question answering, reference sets}
  \vskip 0.3in
]
\printAffiliationsAndNotice{}
% GT-FENG manuscript v16: shared body of the public preprint (main.tex) and the anonymous ICML submission (main_icml.tex).
% Table bodies in tables/*.tex are generated from the result files by scripts/make_v16_tables.py.

\begin{abstract}
Whether a language model's answer is correct often depends on jurisdiction, time, and definition. A wrong answer may match several documented rules, so rule-based error attribution is ambiguous and depends on the evaluator's reference set. We introduce GT-FENG (General Theory of Factual Evaluation and Normative Grounding), a framework that evaluates contextual correctness, retains every matching rule under a stated source mapping, and reports how attribution changes across nested reference sets. We show that answer values alone cannot identify which source a model used when distinct sources share a value, and we propose the FENG value, a measure of rule matching corrected against nearby-value perturbations. In nine models' answers on Medicaid, the share of wrong answers attributed only to other states falls from 58.6\% to 33.1\% as the reference set grows, and 10.3 of these 25.5 percentage points exceed the drop observed when answers are shuffled across states; unemployment insurance and 11 models' released answers on FiscalQA Pro show the same dependence. We check the labels against planted sources of known origin and test the evaluator under a matched-settings rerun and on a Boolean version of the questions. \gtfengrelease
\end{abstract}

\section{Introduction}

People ask language models about rules, whether it is the income limit for a public benefit or the deadline for a tax filing. The correct answer depends on the jurisdiction, date, and definition specified in the question.

Standard factuality evaluations check answers against reference answers or sources \citep{min-etal-2023-factscore,wei2024simpleqa}, but do not show which documented rules match a wrong answer. \citet{feng2026right} report that Claude Sonnet 5.5 gave 41\% of the federal poverty level as Oklahoma's Medicaid income limit for parents in July 2025. The answer is wrong, and it equals Iowa's current standard and two earlier Oklahoma standards. A count of cross-state errors would attribute this answer to Iowa, and a count of outdated answers would attribute it to Oklahoma's past value.

Evaluations of time- and place-dependent questions count such errors, for example outdated answers to statutory questions \citep{prior2026asking,nicolaysen2026norregel} and answers that give another state's value \citep{feng2026right}. These counts depend on two choices: how to count an answer that matches several rules, and which records the evaluator documents. Two evaluations of the same answers can therefore report different counts.

To formalize these evaluation choices, we introduce GT-FENG (General Theory of Factual Evaluation and Normative Grounding), a framework for evaluating contextual and temporal correctness and analyzing rule-based error attribution in rule-governed domains such as law and public policy. For each wrong answer, GT-FENG identifies all matching documented rules. When the matched rules point to different errors, such as another state and an earlier year, GT-FENG labels the answer ambiguous. By expanding the reference set in stages and recounting errors in each category, GT-FENG quantifies how attribution changes with documentation coverage. When distinct sources share a value, answer values alone cannot identify the source used by a model (Section~\ref{sec:theory}). We also propose the FENG value (Section~\ref{sec:feng}), a baseline-corrected metric of how often wrong answers match documented rules.

We evaluate GT-FENG on answers from nine models to questions on Medicaid and state unemployment insurance, and we re-score released answers from 11 models on FiscalQA Pro, a published benchmark of French tax law \citep{cymbler2026temporal}. In Medicaid, the share of wrong answers attributed only to other states falls from 58.6\% to 33.1\% as the reference set grows. Shuffling answers across states shows that 10.3 of these 25.5 percentage points exceed the drop that the denser reference set produces alone and reflect an association between wrong answers and the asked state's own records. In FiscalQA Pro, the share of closed-book wrong answers attributed only to other legal versions rises from 3.1\% to 32.6\% and then falls to 21.4\%. Three controlled analyses test the evaluator: answers to prompts with a planted source of known origin, a rerun under matched settings without the stability filter, and a Boolean version of the Medicaid questions.

Our contributions are:
\begin{itemize}[leftmargin=*,itemsep=1pt,topsep=2pt]
\item A formal framework for contextual correctness and for matching wrong answers to indexed normative rules (Section~\ref{sec:framework}).
\item A separation of shift ambiguity from source ambiguity under an explicit source mapping, and the result that answer values cannot identify a source shared with another (Section~\ref{sec:theory}).
\item Reference-set-dependent attribution over nested reference sets, calibrated against a shuffle null, with a standard report (Sections~\ref{sec:eval} and \ref{sec:results}).
\item The FENG value, its local perturbation baseline, and its behavior on three instances and on planted sources of known origin (Sections~\ref{sec:feng}, \ref{sec:results} and \ref{sec:validation}).
\end{itemize}

\section{Related work}

\textbf{Contextual and temporal factuality.} Factual answers can depend on when and where a question is asked \citep{dhingra-etal-2022-time,zhang-choi-2021-situatedqa}, and benchmark answers themselves can become outdated \citep{jiang2026benchmarks}. In law, correctness depends on jurisdiction and the applicable version of a rule \citep{zheng2025reasoning,dahl2024large}. SituatedQA \citep{zhang-choi-2021-situatedqa} evaluates questions whose answers vary across locations or time, while CrossLex \citep{yang2026crosslex} tests whether models apply the appropriate legal rules across jurisdictions. Recent legal benchmarks focus more specifically on temporal validity. \citet{prior2026asking} examine stale and recency-biased statutory answers, \citet{sobhani2026time} test legal version resolution, and \citet{nicolaysen2026norregel} distinguishes correct, outdated, and other incorrect values against dated statutory histories. FiscalQA Pro \citep{cymbler2026temporal} documents cases in which retrieval systems cite the current version of a law when another version applies. These studies evaluate correctness across contexts, but do not systematically examine which documented rules also match an incorrect answer or how the matches change with the reference set.

\textbf{Factuality evaluation and error attribution.} Existing factuality frameworks primarily evaluate whether generated claims are supported by evidence \citep{rashkin-etal-2023-measuring,iqbal-etal-2024-openfactcheck} or classify the errors in generated responses \citep{xu-etal-2025-diagnosing}. Closely related work on source attribution examines cases where a claim is supported by evidence but attributed to the wrong source; \citet{alvarez2026provenanceguard} study cross-source conflation in tool-using agents. GT-FENG instead examines which documented rules agree with the value of an incorrect answer. When multiple rules share the same value, the answer alone cannot distinguish among them, so GT-FENG retains all matching rules.

\textbf{Normative rules and reference-set dependence.} Legal knowledge systems represent the scope of rules: LegalRuleML \citep{oasis2021legalruleml} encodes jurisdiction, authority, and temporal validity, while rule-governed retrieval systems resolve version and scope before generation \citep{wang2026version}. Other work evaluates normative reasoning when available evidence is inconsistent or defeasible \citep{ulizko2026benchmarking}. These approaches address how rules are represented, selected, or applied, and leave open how the coverage of an evaluation's reference set affects error attribution. \citet{feng2026right} provide an example of this problem: wrong Medicaid answers matched other states' current values as well as the asked state's historical values or alternative reporting conventions, and the apparent cross-state error count changed as additional records were considered. GT-FENG formalizes this observation and extends it: it enumerates matching rules under a source mapping, separates shift from source ambiguity, calibrates the change in exclusive attribution against a shuffle null, introduces a baseline-corrected measure of rule matching, and applies these to unemployment insurance, French tax law, and controlled analyses (Table~\ref{tab:related}).

\begin{table*}[t]
\caption{Comparison with related evaluation methods.}\label{tab:related}
\centering\small\setlength{\tabcolsep}{3pt}
\begin{tabular}{@{}lccccc@{}}
\toprule
\textbf{Work} & \textbf{Contextual} & \textbf{Error-rule} & \textbf{All matches} & \textbf{Nested sets} & \textbf{Baseline} \\
\midrule
SituatedQA \citep{zhang-choi-2021-situatedqa} & Yes & Partly & No & No & No \\
norregel \citep{nicolaysen2026norregel} & Yes & Partly & No & No & No \\
\citet{prior2026asking} & Yes & No & No & No & No \\
FiscalQA Pro \citep{cymbler2026temporal} & Yes & Partly & No & No & No \\
CrossLex \citep{yang2026crosslex} & Yes & Partly & No & No & No \\
OpenFactCheck \citep{iqbal-etal-2024-openfactcheck} & Yes & No & No & No & No \\
\citet{xu-etal-2025-diagnosing} & No & No & No & No & No \\
ProvenanceGuard \citep{alvarez2026provenanceguard} & No & No & No & No & No \\
\citet{feng2026right} & Yes & Yes & No & Yes & No \\
GT-FENG & Yes & Yes & Yes & Yes & Yes \\
\bottomrule
\end{tabular}

\vspace{2pt}{\footnotesize\emph{Note:} Entries indicate features explicitly evaluated or formalized in the cited work. Partly denotes a limited form of error-rule matching. The baseline metric corrects observed rule matches against a specified reference baseline.}
\end{table*}

\section{The GT-FENG framework}\label{sec:framework}

\textbf{Normative facts.} We call a fact whose value depends on the jurisdiction, the date, and the legal definition in use a \emph{normative fact}: its value is set by a rule, and the rule applies only within its context. GT-FENG indexes each rule by four coordinates (Table~\ref{tab:notation}): the quantity $q$, the jurisdiction $j$, the time $t$, and the definition $d$ (the reporting convention or legal definition under which the value is stated). The gold value $G(q,j,t,d)$ is the value that the rule sets at that index. $G$ can return a number (an income limit), a Boolean (whether a duty applies), a category, or a set. Equality is decided by a comparison rule fixed for each domain before scoring: numbers after normalization to the unit and precision of the source, Booleans and categories exactly, and sets as sets. A domain may add coordinates such as authority or covered population.

\textbf{Definition 1 (Contextual correctness).} A question asks for one index $(Q,J,T,D)$. An answer $A$ is \emph{contextually correct} when $A = G(Q,J,T,D)$. Correctness is therefore a relation between an answer and a context: the same $A$ can be correct in one context and wrong in another.

\textbf{Definition 2 (Grounding).} A wrong answer $A$ is \emph{grounded in} the rule at index $(q,j,t,d)$ when that index is documented in the reference set $R$ and $G(q,j,t,d) = A$.

\subsection{Matching and attribution}

\textbf{Definition 3 (Match set).} For a wrong answer, the match set $M$ contains every index $(q,j,t,d) \in R$ with $(q,j,t,d) \neq (Q,J,T,D)$ and $G(q,j,t,d) = A$. If $M$ is empty, the answer is labeled U (unmatched).

\textbf{Definition 4 (Shift).} The shift of a match is the set of coordinates in which it differs from $(Q,J,T,D)$, written as a string of coordinate letters. A match $(Q,j,T,D)$ with $j \neq J$ has shift J; a match $(Q,J,t,D)$ with $t \neq T$ has shift T; a match that differs in both has shift JT.

\textbf{Definition 5 (Source mapping).} For each question $x$, a map $\sigma_x$ assigns every documented index $i$ a source identity $\sigma_x(i)$. The evaluator specifies the map before observing model answers and holds it fixed across reference levels. The default groups indices with equal $(q,j,d)$ across time; applications may instead group by document or legal provision.

\textbf{Definition 6 (Label and ambiguity).} For nonempty $M$, if every match has the same shift S, the answer is labeled S. Otherwise it is labeled \emph{ambiguous}, and the label lists every shift in $M$. GT-FENG distinguishes two kinds of ambiguity:
\begin{itemize}[leftmargin=*,itemsep=1pt,topsep=2pt]
\item \emph{Shift ambiguity}: $M$ contains matches with two or more different shifts. Oklahoma's parent standard for July 2025 is not 41\%. At level R3 (Section~\ref{sec:eval}), 41\% is Iowa's current standard (shift J), Oklahoma's 2018 standard (shift T), and Oklahoma's 2022 standard under the other disregard convention (shift TD), so the label is ambiguous \{J, T, TD\}.
\item \emph{Source ambiguity}: $M$ contains $i_1, i_2$ with $\sigma_x(i_1) \neq \sigma_x(i_2)$. GPT-5.6 Sol answers 305\% for Massachusetts's pregnancy limit (gold 205\%). At R3, 305\% is the current limit of both Missouri and West Virginia. Both matches have shift J, so the label is J and the answer is source-ambiguous.
\end{itemize}
The two kinds are counted separately: a single source can carry two shifts, and two sources can share one shift.

\textbf{Families.} An item is \emph{grounded in} a family of shifts F, an error category, when $M$ is nonempty and every shift in $M$ belongs to F. F can be \{J, JD\} for cross-jurisdiction errors or \{T, TD\} for outdated values. The category is exclusive: a documented match outside F removes an item from it even though its earlier matches remain. We also report the \emph{any-F} share (at least one match in F) and the \emph{contested} share (a match in F and one outside F).

\textbf{Reference-set expansion.} The evaluator reports labels at nested reference sets $R_1 \subseteq R_2 \subseteq \cdots$. $R_1$ contains the indices that a naive check would use, and the final level contains the most complete documentation available.

\subsection{The FENG value}\label{sec:feng}

\textbf{Definition 7 (Local perturbation baseline and FENG value).} For numeric values and a family F, $O$ is the number of wrong items grounded in F and $W$ the number of wrong items. For each wrong item, the evaluator replaces $A$ by every nearby value $A+i$, where $i$ runs over a fixed set of increments $I$. $I$ excludes increments equal to a known definition shift, and increments that reach the correct value are dropped for that item. $N$ is the sum over wrong items of the share of nearby values that are grounded in F, and $E = O - N$. The p-value is the share of $P$ random perturbations, each taking one increment per item, for which the number of grounded items reaches $O$. The FENG value is $(O - N)/(W - N)$, defined when $W > N$.

The formula has the form of Cohen's kappa \citep{cohen1960coefficient}. It is 0 when wrong answers are grounded in F as often as nearby values are, 1 when every wrong answer is grounded in F, and negative when nearby values are grounded in F more often than the answers themselves.

\textbf{Standard report.} At each reference level, a GT-FENG evaluation reports correct items, items labeled U, single-shift labels by shift, shift-ambiguous items, source-ambiguous items, the mean number of distinct sources in $M$ among matched items, $O$ for each family of interest, and the FENG value for numeric values. \gtfengpackage

\section{Theoretical properties}\label{sec:theory}

When the reference set grows, some changes in a GT-FENG report follow from its construction and others reflect the model's answers, and an evaluation needs to tell them apart. Propositions 1 to 3 are elementary consequences of the definitions and give the monotonicity of the match set, the condition under which exclusive attribution cannot increase, and the limit of identifying a source from an answer value. Write $M(R)$ for the match set computed with reference set $R$.

\textbf{Proposition 1 (Monotonicity).} Let $R_1 \subseteq R_2$. For every item, $M(R_1) \subseteq M(R_2)$. Hence (a) an item labeled U under $R_2$ is labeled U under $R_1$, and (b) an item that is shift-ambiguous or source-ambiguous under $R_1$ stays so under $R_2$.

\emph{Proof.} Definition 3 selects indices from $R$ by a condition that does not depend on $R$, so a larger $R$ yields a larger or equal $M$. For (a), $M(R_2)$ empty implies $M(R_1)$ empty. For (b), two matches in $M(R_1)$ also belong to $M(R_2)$, with their shifts and fixed source identities unchanged. $\square$

\textbf{Proposition 2 (Upper bound under coverage).} Let F be a family of shifts and $R_1 \subseteq R_2$. Assume coverage: $R_1$ contains every index of $R_2$ whose shift belongs to F. Then every item grounded in F under $R_2$ is grounded in F under $R_1$, so $O$ cannot increase from $R_1$ to $R_2$.

\emph{Proof.} Take an item grounded in F under $R_2$. Every match in $M(R_2)$ has a shift in F, so by coverage its index lies in $R_1$, and Definition 3 places it in $M(R_1)$. Hence $M(R_2) \subseteq M(R_1)$, and with Proposition 1, $M(R_1) = M(R_2)$. $\square$

A larger reference set increases match coverage, but it need not increase exclusive attribution. When the target family is not yet fully covered, new matches within F can add items to $O$; once records outside F enter, items already in $O$ can become ambiguous. FiscalQA Pro shows both phases (Section~\ref{sec:results}). The any-F share never falls as $R$ grows, by Proposition 1, and the direction of the exclusive share follows from the reference sets; whether the size of its change reflects structure in the answers requires a null distribution (Section~\ref{sec:eval}).

\textbf{Proposition 3 (Attribution non-identifiability).} Suppose a wrong answer has matches $i_1, i_2 \in M$ with $\sigma_x(i_1) \neq \sigma_x(i_2)$. Consider two accounts: the model drew on source $\sigma_x(i_1)$, or on source $\sigma_x(i_2)$. Any attribution procedure that uses only the question, the answer, and $R$ returns the same output under both accounts, so a procedure that names a single source is wrong under at least one of them.

\emph{Proof.} Both sources set the value $A$, so both accounts produce the same observable answer. The question, $R$, and the prespecified source mapping are fixed, so a procedure that is a function of these observations returns the same output under both accounts. If that output names one source, it is wrong under the other account. $\square$

GT-FENG therefore reports the documented candidates and their source identities. Attribution heuristics such as checking the asked jurisdiction first add an assumption about sources, and with it they can select the right source more often. The quantity $1/k$ is the probability that a uniform choice among the $k$ matching sources is right when the true source is one of them; it is the baseline for an attributor that uses no information beyond $M$. Individual sources can be unidentifiable while an association between wrong answers and the asked jurisdiction across many answers remains detectable, and Section~\ref{sec:eval} tests it by shuffling answers across jurisdictions.

\section{Instances and evaluation design}\label{sec:eval}

\textbf{Medicaid.} We instantiate GT-FENG on the Medicaid eligibility data of \citet{feng2026right}. Q is one of three quantities (pregnancy, infant, parent). J is one of 51 jurisdictions (the 50 states and the District of Columbia). T is July 2025, and earlier times are the MACStats editions 2018 to 2024. D is the treatment of the 5-point income disregard (included or excluded). Answers are rounded to whole percentage points, the precision of MACStats. Federal anchors (100\%, 133\%, 138\%) form a separate fixed set; a match to an anchor carries the label federal, a match to another number in the asked state's data book carries the label Q, and both lie outside F = \{J, JD\}. The source mapping groups the asked state's data-book values under one source identity and uses the default $(q,j,d)$ grouping for other state records. $R_1$ (Table~\ref{tab:levels}) lists every index with shift J or JD, so the coverage assumption of Proposition 2 holds for F = \{J, JD\}.

\begin{table}[t]
\caption{Reference levels for Medicaid.}\label{tab:levels}
\centering\small\setlength{\tabcolsep}{3pt}
\begin{tabular}{@{}lp{0.62\columnwidth}l@{}}
\toprule
\textbf{Level} & \textbf{Indices added} & \textbf{Shifts} \\
\midrule
R1 & other states, July 2025, both definitions & J, JD \\
R2 & asked state: other definition; editions 2021, 2023, 2024; federal anchors & D, T, TD \\
R3 & asked state: every edition 2018 to 2024 & T, TD \\
R4 & asked state: every other number in its data-book rows 2018 to 2025 & Q \\
\bottomrule
\end{tabular}
\end{table}

\textbf{Unemployment insurance (UI).} The values come from the U.S.\ Department of Labor's Significant Provisions of State Unemployment Insurance Laws, January 2018 to January 2026. Q is the maximum weekly benefit amount, the minimum weekly benefit amount, or the taxable wage base; T is January 2026; D separates amounts with and without dependents' allowances, or the standard rate group from an alternative one. The UI reference levels (Table~\ref{tab:uilevels}) parallel those of Medicaid.

\textbf{Models.} Nine models answer each of the 153 questions per program twice: Claude Sonnet 5.5, GPT-5.6 Sol, and seven open-weight models (Table~\ref{tab:settings}). An item is \emph{stable}, and enters the main analysis, when both answers agree.

\textbf{FiscalQA Pro.} This benchmark \citep{cymbler2026temporal} asks for amounts set by the French General Tax Code at stated dates. Its authors released a versioned corpus of the code, the answers of 11 models, and the scoring keys. We apply GT-FENG to the released answers to the 209 scored temporal questions, with no new model runs. Q is one of 33 benchmark families, each an amount set by one article, and T is the date in the question; France is the only jurisdiction, and no definition coordinate is used. A response is correct when any amount it states is within 1 of the gold value, the benchmark's own numeric rule. For a wrong response, $A$ is the first amount it states, with its unit (euro, franc, or percent). Amounts match only within the same unit. R1 holds the family's gold values at the other benchmark dates (shift T), R2 adds every amount in every other version of the article (shift QT), and R3 adds the other amounts of the applicable version (shift Q); F = \{T, QT\} collects values from another time. Nearby values change the answer by 1\% to 4\%. We report condition A (no retrieval) and condition B (retrieval from a corpus of current versions), out of four conditions (Table~\ref{tab:fiscalcond}).

\textbf{Uncertainty and nulls.} FENG confidence intervals are 95\% percentile intervals over 10,000 bootstrap samples of whole jurisdictions (families for FiscalQA Pro) \citep{efron1979bootstrap}, with the same samples at every level. To separate the drop expected from a denser reference set from an association with the asked state, we shuffle each model's stable answers across jurisdictions within each quantity and re-evaluate every level \citep{ernst2004permutation}. The \emph{excess drop} is the observed drop of the exclusive share from R1 to R4 minus its mean under 2,000 shuffles; its 95\% interval resamples jurisdictions with the shuffled mean held fixed. A second permutation test compares the share of wrong answers that match a record of the asked state (a match with shift D, T, TD, or Q) with the shuffled share. P-values over models are Holm-adjusted within each test and program \citep{holm1979simple}.

\section{Results}\label{sec:results}

\begin{table*}[t]
\caption{Medicaid results for nine models (stable items). O counts wrong answers grounded only in F = \{J, JD\}.}\label{tab:medicaid}
\centering\small\setlength{\tabcolsep}{3pt}
\begin{tabular}{@{}lcccccccc@{}}
\toprule
\textbf{Model} & \textbf{Accuracy} & \textbf{Top-answer share} & \textbf{W} & \textbf{O at R1} & \textbf{O at R4} & \textbf{Source-amb.\ at R4} & \textbf{FENG at R1} & \textbf{FENG at R4} \\
\midrule
GPT-5.6 Sol & 52.1\% & 7.7\% & 68 & 47 & 15 & 44 & 0.50 & $-$0.10 \\
Claude Sonnet 5.5 & 47.5\% & 13.6\% & 62 & 37 & 8 & 42 & 0.28 & $-$0.27 \\
DeepSeek V4 Flash & 25.2\% & 10.7\% & 98 & 60 & 18 & 66 & 0.22 & $-$0.40 \\
Llama 3.3 70B & 18.6\% & 17.8\% & 96 & 66 & 36 & 60 & 0.34 & $-$0.05 \\
Mistral Small 3.2 24B & 6.9\% & 41.4\% & 108 & 74 & 34 & 72 & 0.42 & $-$0.16 \\
Gemma 4 31B & 3.0\% & 72.6\% & 131 & 82 & 29 & 79 & 0.34 & $-$0.31 \\
Qwen3.6 35B-A3B & 2.8\% & 53.8\% & 141 & 109 & 100 & 57 & 0.49 & 0.43 \\
Llama 3.1 8B & 2.6\% & 74.5\% & 149 & 100 & 85 & 99 & 0.13 & $-$0.005 \\
Gemma 3 12B & 0.0\% & 54.7\% & 128 & 0 & 0 & 0 & $-$0.02 & $-$0.02 \\
\bottomrule
\end{tabular}
\end{table*}

\textbf{Reference-set dependence.} Documenting the asked state's own records turns many apparent cross-state errors into ambiguous ones (Table~\ref{tab:medicaid}). Summed over the nine models, O falls from 575 to 325, and the exclusive share of the 981 stable wrong answers falls from 58.6\% to 33.1\%. The any-F share stays at 58.6\% at every level, so every answer that leaves O keeps its cross-state matches and gains a match outside F. At R4, more than half of the stable wrong answers are source-ambiguous for seven of the nine models, so by Proposition 3 no single source can be identified for them. The counts O for Claude and GPT reproduce the four protocol levels of \citet{feng2026right} exactly. In UI, O falls from 239 to 173, and the exclusive share from 27.5\% to 19.9\% (Table~\ref{tab:uimain}).

\textbf{Drop beyond chance.} A falling exclusive share does not by itself show that wrong answers match the asked state's records more often than chance, because added records lower exclusive attribution mechanically. The shuffle null of Section~\ref{sec:eval} keeps each model's answers and the reference sets and breaks only the pairing of answers with jurisdictions. In Medicaid, the exclusive share falls by 25.48 points from R1 to R4, against 15.15 points under shuffling: an excess drop of 10.33 points (95\% CI [6.98, 13.79], $p < 0.001$; Table~\ref{tab:excess}), so about 60\% of the drop comes from the denser reference set and about 40\% from the answers.

The excess is concentrated in four models: Claude (28.1 points), DeepSeek (29.2), GPT (32.2), and Llama 3.3 (20.1), each with an interval above 0 and Holm $p = 0.004$. Mistral's excess is 8.1 points (Holm $p = 0.005$), and Gemma 4, Qwen3.6, Llama 3.1, and Gemma 3 show no excess. In unemployment insurance, the pooled excess is 3.7 points [1.0, 6.5] (Table~\ref{tab:excessui}).

\begin{table}[t]
\caption{Medicaid exclusive-share drop from R1 to R4 (points), observed and under 2,000 shuffles; values are rounded separately.}\label{tab:excess}
\centering\scriptsize\setlength{\tabcolsep}{2.5pt}
\begin{tabular}{@{}lcccc@{}}
\toprule
\textbf{Model} & \textbf{Drop} & \textbf{Shuffled} & \textbf{Excess [95\% CI]} & \textbf{Holm p} \\
\midrule
GPT-5.6 Sol & 47.1 & 14.8 & 32.2 [21.4, 43.4] & 0.004 \\
Claude Sonnet 5.5 & 46.8 & 18.7 & 28.1 [14.2, 42.3] & 0.004 \\
DeepSeek V4 Flash & 42.9 & 13.7 & 29.2 [19.3, 39.5] & 0.004 \\
Llama 3.3 70B & 31.2 & 11.1 & 20.1 [11.0, 28.9] & 0.004 \\
Mistral Small 3.2 24B & 37.0 & 29.0 & 8.1 [$-$0.8, 16.7] & 0.005 \\
Gemma 4 31B & 40.5 & 38.5 & 2.0 [$-$5.1, 8.5] & 0.372 \\
Qwen3.6 35B-A3B & 6.4 & 5.4 & 1.0 [$-$2.6, 5.2] & 0.446 \\
Llama 3.1 8B & 10.1 & 8.8 & 1.3 [$-$3.5, 6.9] & 0.438 \\
Gemma 3 12B & 0.0 & 0.0 & 0.0 [0.0, 0.0] & 1.000 \\
Pooled & 25.5 & 15.1 & 10.3 [7.0, 13.8] & -- \\
\bottomrule
\end{tabular}

\end{table}

\textbf{Association with the asked state's records.} The second permutation test gives the same grouping. At R4, 77.4\% of Claude's stable wrong answers match a record of the asked state, against 19.0\% after shuffling; the differences are 50.7, 53.8, and 34.3 points for GPT, DeepSeek, and Llama 3.3 (Holm $p < 0.001$; Table~\ref{tab:perm_med}). For Mistral the difference is 10.0 points (39.8\% against 29.8\%), and Gemma 4, Qwen3.6, Llama 3.1, and Gemma 3 do not differ from their shuffled shares after adjustment. Of the 250 answers that leave O between R1 and R4, 244 match a record of the asked state. In UI, the share exceeds the shuffled share for all nine models, by 3.5 to 65.8 points (Holm $p \leq 0.007$; Table~\ref{tab:uiperm}).

\begin{figure*}[t]\centering
\includegraphics[width=0.86\textwidth]{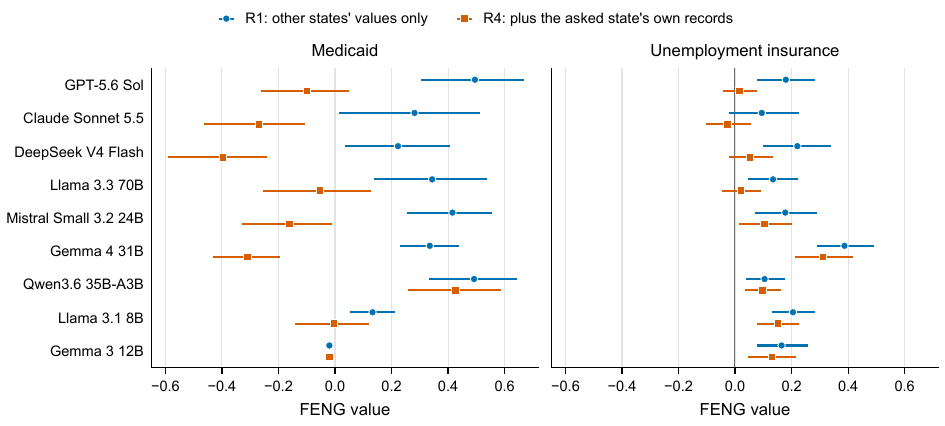}
\caption{FENG value at R1 and R4 with 95\% confidence intervals, Medicaid and unemployment insurance. The Medicaid interval of Gemma 3 12B, [$-$0.03, $-$0.01], is narrower than its markers.}\label{fig:feng}
\end{figure*}

\textbf{FENG value.} At R1, eight of the nine Medicaid models give wrong answers that match other states' values more often than slightly different numbers would, with 95\% intervals above 0 (Figure~\ref{fig:feng}). At R4, the value is lower for each of these eight models and negative for seven of them; for seven, the interval of the drop excludes 0. A negative value means that wrong answers gain matches to the asked state's records more often than nearby numbers do, the association that the permutation test measures. Because $N$ also changes with $R$, the propositions do not fix the direction of the FENG value, and the drop is an empirical result. Accuracy and reliance on one default value do not determine it: Gemma 4 and Qwen3.6 both answer about 3\% of items correctly and repeat one value across many states, and their R4 values have opposite signs, $-$0.31 and 0.43 (Appendix~\ref{app:medicaid}). In UI, the R1 value is positive for all nine models and lower at R4 for all nine.

\textbf{FiscalQA Pro.} The share of wrong answers attributed to another time depends on which versions are documented (Table~\ref{tab:fiscal}). With only the benchmark's own question dates documented, 3.1\% of the closed-book wrong answers match only other dates; with every version of the article documented, 32.6\% match only other versions. For retrieval over current versions, the share rises from 1.0\% to 79.5\%. R1 omits most temporal matches within F, so Proposition 2 does not apply to the first expansion; from R2 to R3, new matches in the applicable version, outside F, lower the shares to 21.4\% and 65.6\%. The FENG value at R2 is positive with an interval above 0 for all 11 models in both conditions, and it is lower at R3 for all 22 model-condition pairs (Table~\ref{tab:fiscalfeng}). Wrong answers match amounts of the asked article more often than shuffled answers do: closed-book, 14.0\% to 60.7\% of wrong answers equal an amount of some version of it, against 2.3\% to 7.6\% after shuffling across questions; under retrieval, 59.4\% to 91.2\% against 6.7\% to 11.5\% (Holm $p \leq 0.0011$; Table~\ref{tab:fiscalperm}). Under that retrieval, 59.9\% of the wrong answers equal an amount of the version in force today that the applicable version does not contain, and 72.8\% are amounts of a retrieved version's text (Table~\ref{tab:retrieval}). The remaining wrong B answers have a FENG value of 0.24 at R2, close to the closed-book 0.27, so copying from retrieval and confusion between versions both appear (Appendix~\ref{app:fiscal}).

\begin{table*}[t]
\caption{FiscalQA Pro across 11 models: wrong answers grounded only in other versions (F = \{T, QT\}), and the range of FENG values over models.}\label{tab:fiscal}
\centering\small\setlength{\tabcolsep}{4pt}
\begin{tabular}{@{}lccccccc@{}}
\toprule
\textbf{Condition} & \textbf{W} & \textbf{O at R1} & \textbf{O at R2} & \textbf{O at R3} & \textbf{Shift-amb.\ R3} & \textbf{FENG R2} & \textbf{FENG R3} \\
\midrule
A, closed-book & 1,952 & 61 (3.1\%) & 637 (32.6\%) & 418 (21.4\%) & 219 & 0.07 to 0.58 & 0.02 to 0.45 \\
B, current-version retrieval & 2,027 & 21 (1.0\%) & 1,612 (79.5\%) & 1,329 (65.6\%) & 283 & 0.58 to 0.91 & 0.47 to 0.85 \\
\bottomrule
\end{tabular}

\end{table*}

\section{Controlled validation and robustness}\label{sec:validation}

The planted-source, rerun, and Boolean analyses use the seven open-weight models at fixed providers (designs in Appendices~\ref{app:planted} to \ref{app:boolean}).

\textbf{Planted sources.} To test how the reference set affects the labels of an answer whose source is known, we fix the source by design: we prepend to each Medicaid question one reference entry that names a state, the quantity, a value X, and the disregard convention (Appendix~\ref{app:planted}). In condition PJ, X is the July 2025 value of a random other state (153 questions); in condition PT, X is the asked state's own 2018 value (38 questions where it differs from the gold value), so the two conditions test complete and incomplete coverage of the true source. Of 1,337 answers at temperature 0, 1,206 (90.2\%) give the planted value, and we score these.

In PJ, the true source is in $R$ from R1 on, and documenting other records lowers exclusive correct attribution: the share labeled exclusively in the true family falls from 100\% at R1 to 89.4\% at R4, and the FENG value from 0.90 to 0.73 (Table~\ref{tab:planted}). In PT, the true source enters only at R3. At R1, half of the answers are unmatched and the other half are labeled cross-state; at R3, coverage is complete and 49.4\% are labeled exclusively in the true family, the rest ambiguous.

\begin{table}[t]
\caption{Planted sources at R1 and R4, answers that follow the plant (Table~\ref{tab:plantedfull}: all levels). Cov.: true source in $M$; Excl.\ true: only true-family shifts; Unif./Prec.: accuracy of uniform and asked-state-first choice.}\label{tab:planted}
\centering\scriptsize\setlength{\tabcolsep}{1.6pt}
\begin{tabular}{@{}llccccccc@{}}
\toprule
\textbf{Plant} & \textbf{Level} & \textbf{Cov.} & \textbf{Excl.\ true} & \textbf{Shift-amb.} & \textbf{U} & \textbf{Unif.} & \textbf{Prec.} & \textbf{FENG [95\% CI]} \\
\midrule
PJ & R1 & 100\% & 100.0\% & 35.4\% & 0.0\% & 59\% & 59\% & 0.90 [0.88, 0.92] \\
 & R4 & 100\% & 89.4\% & 39.9\% & 0.0\% & 58\% & 57\% & 0.73 [0.65, 0.80] \\
\midrule
PT & R1 & 0\% & 0.0\% & 26.3\% & 49.4\% & 0\% & 0\% & 0.00 [0.00, 0.00] \\
 & R4 & 100\% & 32.7\% & 72.9\% & 0.0\% & 66\% & 100\% & 0.21 [0.07, 0.36] \\
\bottomrule
\end{tabular}

\end{table}

Uniform choice among the matching states is right for 59\% of PJ answers at R1, because 37\% of them match one state and the rest two to six (Table~\ref{tab:plantedk}). Choosing the asked state first is right for all PT answers at R3 and R4 and for 57\% of PJ answers at R4, so its accuracy depends on the mix of sources and rises with the share of PT answers (Table~\ref{tab:plantedmix}).

\textbf{Matched-settings rerun.} The main analysis uses answers that are stable under each model's own serving settings; the rerun checks the results against both choices. We reran the seven open-weight models on all 153 items per program at temperature 0.7 with reasoning off, five repeats per item, and took the modal answer (the most frequent of the five; ties go to the first). Over the same seven models, the Medicaid exclusive share falls from 56.0\% to 35.9\%, against 57.7\% to 35.5\% in the main analysis; in UI it falls from 22.3\% to 17.1\%, against 28.1\% to 21.5\%.

The pooled Medicaid excess drop is 5.8 points [3.0, 8.7], and the R1 FENG value is positive with Holm $p < 0.01$ for six of seven models in each program. Among the models with an excess in the main analysis, DeepSeek (21.5 points) and Llama 3.3 (17.3) replicate (Table~\ref{tab:rerun}).

\textbf{Boolean instance.} To shrink the answer space to two values, we converted the Medicaid questions into yes/no questions about whether a limit exceeds a threshold, with two thresholds per quantity and 39\% to 65\% of states above each, and asked the seven open-weight models twice (2,087 stable items). Accuracy is 56.4\% (95\% CI [53.4, 59.4]), and all 910 stable wrong answers are source-ambiguous, with 46.6 to 50.6 matching source identities per model on average, against 2.0 to 6.2 in the numeric instance at R4 (Table~\ref{tab:boolean}).

\textbf{Design choices.} Under four source mappings, from one source per rule to one per jurisdiction, 73\% to 78\% of matched wrong Medicaid answers are source-ambiguous at R4, and in UI and FiscalQA Pro the share ranges from 38\% to 71\% and from 0\% to 74\% (Table~\ref{tab:sigma}). In all six orders of entry of the three added Medicaid blocks, the excess drop is 10.3 points, and each block contributes 2.6 to 4.1 points on average (Table~\ref{tab:orders}).

Across increment sets from $\pm$1 to $\pm$10, the R1 FENG value keeps its sign for eight of nine Medicaid models and all nine UI models. A resampled-answer null that replaces each wrong answer by the same model's other wrong answers gives values near 0 at R1, so a positive local FENG value shows that wrong answers are real rule values. In synthetic answers, substituted and default values both give a local FENG value of 1 at R1 and no excess drop, while stale own-state values give an excess drop of 0.50, above the 0.20 to 0.32 of the four strong models.

When the first answer of all 153 items is scored, or only the 38 items stable for every model, the pooled Medicaid exclusive share falls by 24.5 and 29.5 points, and the R1 FENG value stays positive for eight and seven of the nine models (Table~\ref{tab:subsets}). In FiscalQA Pro, the first, last, or nearest stated amount leads to the same conclusions (Table~\ref{tab:extraction}).

\section{Conclusion}

When distinct sources share a value, the answer alone cannot identify which source a model used, and the attribution an evaluator reports depends on its reference set and source mapping. In nine models' answers about Medicaid and state unemployment insurance, expanding the reference set reduces the number of wrong answers matching only other states' values from 575 to 325 and from 239 to 173. Calibrated against shuffled answers, about 40\% of the Medicaid drop reflects an association between wrong answers and the asked state's own records, concentrated in four models. In FiscalQA Pro under current-version retrieval, the share of wrong answers matching only other versions of the law rises from 1.0\% at R1 to 79.5\% at R2 and falls to 65.6\% at R3. Rule-based factuality evaluations can report attribution ambiguity, reference-set coverage, the source mapping, and a chance-calibrated change alongside accuracy.

\section*{Limitations}

\textbf{Measurement.} The matches of GT-FENG depend on the coverage and correctness of the documented rules. A missing rule leaves an answer unmatched or gives it the wrong family, as the planted outdated values at R1 show. The reference levels, families of shifts, source mapping, and increments are set by the evaluator, so the counts, ambiguity measures, and FENG values are conditional on them. The interval of the excess drop holds the shuffled mean fixed, so it is narrower than an interval that also resamples the null; for Mistral it includes 0 while the Holm-adjusted shuffle p-value is 0.005.

\textbf{Identification.} A documented match does not show that a model used that source. The local FENG value measures sensitivity to nearby numbers; it does not model how errors are generated or separate substitution from wrong default values that a model repeats. Source counts count the identities defined by $\sigma$, and one document can hold several of them. The jurisdiction shuffle assumes exchangeability of answers within each quantity, and the planted-source test places its sources in the context, so it does not cover sources recalled from training.

\textbf{Scope.} The numeric instances cover three quantities each in Medicaid and unemployment insurance and answers stable across two runs, and the rerun changes several settings at once and covers only open-weight models. In FiscalQA Pro, a wrong response is scored by its first stated amount, and an amount in an article's text is matched without identifying the provision that sets it; documented values are taken as the rules in force without assessing their legal authority.

\section*{Impact Statement}

GT-FENG supports the evaluation of wrong answers from AI systems that answer questions about laws and public benefit programs. Reporting every matching rule and the sensitivity of attribution to the reference set can reduce overconfident attributions of model errors and help auditors interpret evaluation results. The procedure does not replace legal judgment or formal eligibility determination, and its reliability depends on the rule documentation supplied.

\bibliography{refs}
\bibliographystyle{icml2026}

\clearpage
\appendix
\onecolumn
% Appendices follow in a later version.

\end{document}